\documentclass[10pt,twocolumn,letterpaper]{article}

\usepackage[pagenumbers]{iccv}      

\definecolor{iccvblue}{rgb}{0.21,0.49,0.74}
\usepackage[pagebackref,breaklinks,colorlinks,allcolors=iccvblue]{hyperref}
\usepackage{tabularx} 
\usepackage{booktabs} 
\usepackage{float}
\usepackage{algorithmic}
\usepackage{algorithm}
\usepackage{colortbl}

\def\paperID{1024} 
\def\confName{ICCV}
\def\confYear{2025}

\title{AttenST: A Training-Free Attention-Driven Style Transfer Framework with Pre-Trained Diffusion Models}

\author{
Bo Huang\textsuperscript{1}, 
Wenlun Xu\textsuperscript{2}, 
Qizhuo Han\textsuperscript{1}, 
Haodong Jing\textsuperscript{3}, 
Ying Li\textsuperscript{1}\thanks{Corresponding author: lybyp@nwpu.edu.cn} \\
\textsuperscript{1}Northwestern Polytechnical University, 
\textsuperscript{2}Northwest A\&F University, 
\textsuperscript{3}Xi'an Jiaotong University \\
{\tt\small \{bohuang, hqz\}@mail.nwpu.edu.cn, wenlunxu@nwafu.edu.cn, jinghd@stu.xjtu.edu.cn}
}

\begin{document}
\maketitle
\begin{abstract}
While diffusion models have achieved remarkable progress in style transfer tasks, existing methods typically rely on fine-tuning or optimizing pre-trained models during inference, leading to high computational costs and challenges in balancing content preservation with style integration. To address these limitations, we introduce AttenST, a training-free attention-driven style transfer framework. Specifically, we propose a style-guided self-attention mechanism that conditions self-attention on the reference style by retaining the query of the content image while substituting its key and value with those from the style image, enabling effective style feature integration. To mitigate style information loss during inversion, we introduce a style-preserving inversion strategy that refines inversion accuracy through multiple resampling steps. Additionally, we propose a content-aware adaptive instance normalization, which integrates content statistics into the normalization process to optimize style fusion while mitigating the content degradation. Furthermore, we introduce a dual-feature cross-attention mechanism to fuse content and style features, ensuring a harmonious synthesis of structural fidelity and stylistic expression. Extensive experiments demonstrate that AttenST outperforms existing methods, achieving state-of-the-art performance in style transfer dataset.

\end{abstract}

\section{Introduction}
\label{sec:intro}

Style transfer aims to synthesize visually appealing images by merging the content of one image with the artistic style of another. While conventional approaches leveraging convolutional neural networks (CNNs) \cite{STNeural,johnson2016perceptual,zhang2020unified,huang2017AdaIN,liu2021adaattn} and generative adversarial networks (GANs) \cite{karras2021styleGAN,kotovenko2019content,zhu2017unpaired,CartoonGAN,isola2017pix2pix} have achieved notable success, they frequently encounter limitations in terms of flexibility, generalization capability, and style diversity. The emergence of diffusion models has revolutionized generative tasks, establishing new benchmarks in image generation \cite{ramesh2022hierarchical,dhariwal2021diffusion,rombach2022high}, super-resolution \cite{li2022srdiff}, and image editing \cite{kawar2023imagic}. This innovative paradigm has recently been extended to style transfer, demonstrating remarkable potential through various implementations \cite{zhang2023InST,chung2024styleID,everaert2023diffusionstyle,yang2023zero,wang2023stylediffusion,hertz2024styleAlign,jeong2024DiffStyle}.

Previously, diffusion-based style transfer methods \cite{zhang2023InST,ramesh2022hierarchical,li2023stylediffusion2,li2023moecontroller,ruiz2023dreambooth,qi2024deadiff} predominantly rely on fine-tuning  pre-trained models or optimizing the inference stage \cite{wang2024instantstyleplus,liu2023more}. However, such methods necessitate significant computational resources and data, with limited adaptability. The advent of training-free methods \cite{jeong2024DiffStyle,chung2024styleID,hertz2024styleAlign,wang2024instantstyleplus} has shed light on these issues, yet existing approaches still face some challenges: 

\textbf{Style Injection.} 
Existing methods \cite{hertz2024styleAlign,jeong2024visual,tumanyan2023plug,chung2024styleID} have not systematically explored style injection at different feature levels, leading to inadequate style transfer at critical feature layers and ultimately yielding suboptimal style representation in generated images.

\textbf{Content Preservation.} Excessive style injection can result in the degradation of fine details and structural integrity. Training-free methods typically employ ControlNet \cite{zhang2023controlnet} to constrain the content of generated images. However, it relies on additional input data that may inadequately capture fine details of the content image, especially in complex scenes, resulting in content degradation in the generated images.

Recent efforts have studied the manipulation of attention mechanisms in diffusion models to facilitate personalized image generation. Prompt-to-Prompt \cite{hertz2023prompt}, a pioneering work in attention modification, manipulates cross-attention layers to control the spatial relationship between image layouts and textual prompts. Similarly, Plug-and-Play \cite{tumanyan2023plug} refines structural and layout control by manipulating spatial features. However, these methods often struggle to accurately incorporate the desired features and are prone to distorting the original image.

Drawing inspiration from cutting-edge developments in attention-based image generation, we present AttenST, an attention-driven framework for style transfer. Recognizing that the $query$ in self-attention encodes semantic information and spatial layout, while $key$ and $value$ determine rendered attributes and elements, we introduce a style-guided self-attention mechanism. This approach treats style features analogously to text conditioning in cross-attention, providing explicit guidance for the generation process. Specifically, our method preserves the content image's $query$ while substituting its $key$ and $value$ with style-derived counterparts, effectively aligning stylistic attributes with content structure while maintaining spatial coherence. Through in-depth analysis of the SDXL architecture, we identify the $5^{th}$ and $6^{th}$ transformer blocks in the decoder as critical for style representation, strategically implementing our style-guided self-attention mechanism at these layers to optimize style transfer quality.

Existing inversion-based style transfer methods \cite{zhang2023InST,tumanyan2023plug,yang2023zero} often fail to address style feature degradation during the inversion process, leading to substantial quality deterioration. To this end, we propose style-preserving inversion (SPI) strategy, which iteratively refines the inversion process at each step. Through multiple resampling iterations, SPI effectively compensates for accumulated errors caused by linear assumptions, yielding more precise inversion sampling points while minimizing style information loss. Our approach generates latent noise representations for both content and style images through this enhanced inversion process. Crucially, the latent representation of the content image serves as the initial noise input for the diffusion model, thereby preserving its essential textural and structural characteristics throughout the style transfer process.


Extensive research \cite{xu2025stylessp,koo2024flexiedit} has demonstrated that precise initialization noise control substantially enhances generation quality. Building upon these insights, we implement adaptive instance normalization (AdaIN) to modulate the content image's latent noise representation by aligning its statistical properties (mean and variance) with style features, enabling early-stage style integration. Nevertheless, this approach tends to compromises content fidelity. To overcome this challenge, we introduce Content-Aware AdaIN (CA-AdaIN), an enhanced normalization technique that integrates content statistics during denoising initialization and employs dual modulation parameters ($\alpha_s$ for style intensity and $\alpha_c$ for content preservation) to achieve optimal balance between stylistic expression and content integrity.
 
To optimize the style-content equilibrium, we introduce a dual-feature cross-attention (DF-CA) mechanism that leverages image encoders to extract feature representations from both style and content inputs, while precisely controlling the generation process through cross-attention modulation.

Our contributions can be summarized as follows:
\begin{itemize}
    \item We propose a style-guided attention mechanism that achieves efficient style feature integration through key-value substitution while maintaining computational efficiency.
    \item We introduce a style-preserving inversion strategy, an iterative refinement process that minimizes style information loss during inversion.
    \item We propose CA-AdaIN, which effectively incorporates style information at the early stage of stylization while mitigating content degradation.
    \item We further optimize the style-content trade-off through the proposed dual-feature cross-attention mechanism, which regulates the generation process by effectively fusing content and style features.
\end{itemize}


\section{Related Works}
\subsection{Neural Style Transfer}

\begin{figure*}[htbp]
    \centering
    \includegraphics[width=1\linewidth]{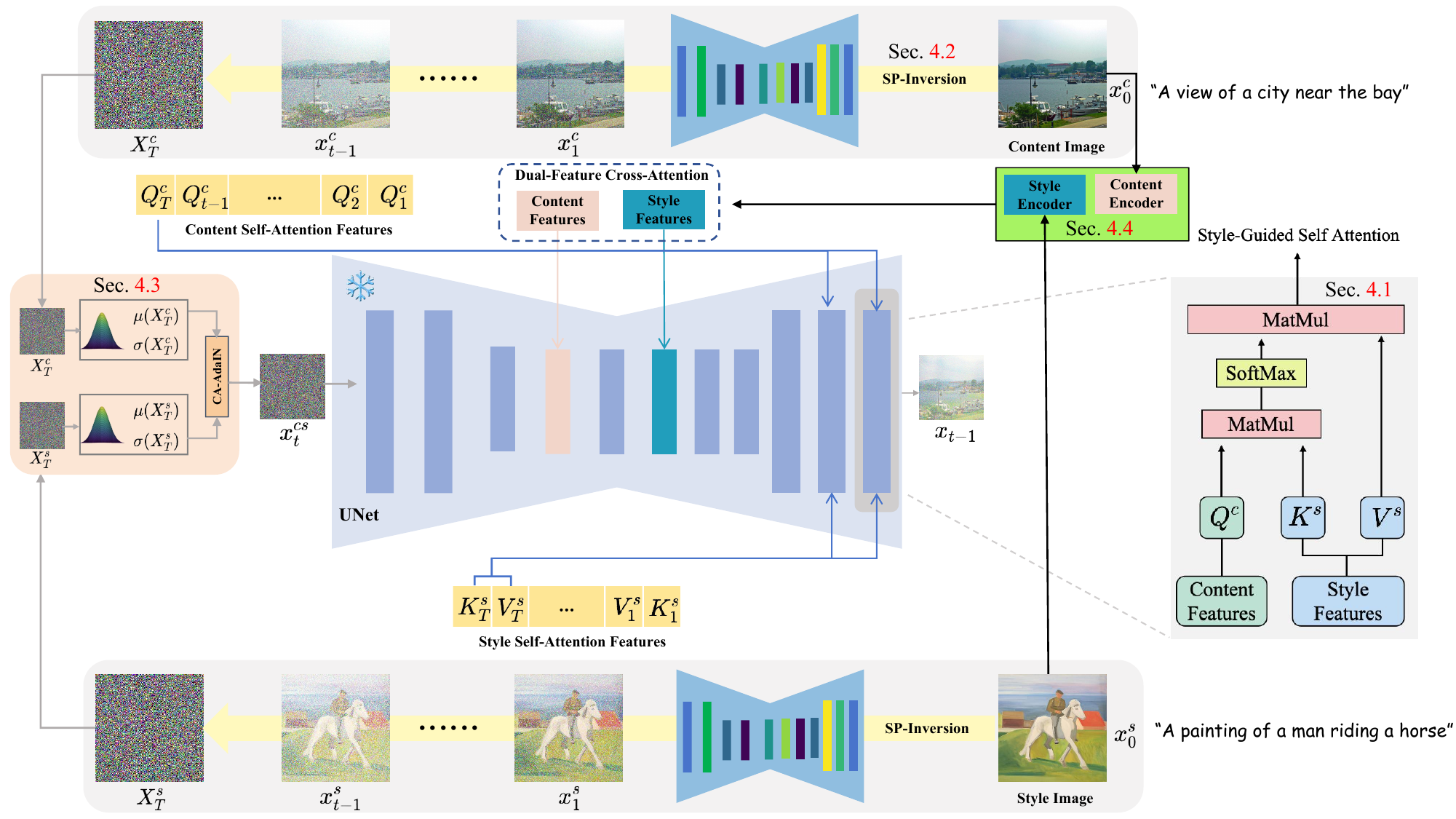}
    \caption{\textbf{Pipeline of the AttenST}. We start with the style-preserving inversion (\cref{4.2}) to invert content image $x^c_0$ and style image $x^s_0$, obtaining their respective latent noise representations, denoted as $X^c_T$ and $X^s_T$. During this process, the query of the content image $Q^{c}$ and the key-value pairs of the style image $(K^{s},V^{s})$ are extracted. Subsequently, the proposed CA-AdaIN mechanism (\cref{4.3}) is employed to refine the latent representation of the content, producing $x^{cs}_t$, which serves as the initial noise input for the UNet denoising process. Throughout denoising, the key and value derived from the self-attention of the style image are injected into the designated self-attention layers (\cref{4.1}), facilitating the integration of style features. Simultaneously, the features of the style and content images are processed through the DF-CA (\cref{4.4}) and incorporated into the corresponding blocks via cross-attention. This strategy constrains the generation process, ensuring effective style integration while preserving the original content, thereby achieving an optimal balance between style and content fidelity.}
    \label{fig:framework}
\end{figure*}

Neural style transfer (NST) leverage neural networks to generate stylized images. Gatys et al. \cite{gatys2016NST} pioneered a style transfer framework using the VGG network \cite{simonyan2014VGG}, where high-level features represent content and low-level features encode style, achieving remarkable results. Nevertheless, this approach requires iterative optimization processes, resulting in substantial computational overhead.

To overcome the computational inefficiency of optimization-based methods. Johnson \etal \cite{johnson2016perceptual} introduced a perceptual loss-driven generative network enabling real-time style transfer in a single forward pass. Huang and Belongie \cite{huang2017AdaIN} further enhanced adaptability and quality by proposing AdaIN, which aligns style and content features through statistical adjustments. More recently, Deng \etal \cite{deng2018aesthetic} introduced an adversarial learning-based framework for style transfer. However, these methods exhibit limited style transfer effectiveness and instability during training.

\subsection{Diffusion Models for Style Transfer}
Diffusion-based NST methodologies have revolutionized the field through their unprecedented precision in style manipulation. Among these breakthroughs, StyleDiffusion \cite{wang2023stylediffusion} presents a novel framework that implements an explicit content representation, coupled with an implicit style learning approach. Parallel developments include InST \cite{zhang2023InST}, which employs an image-to-text inversion paradigm that encodes artistic styles into learnable textual embeddings. DEADiff \cite{qi2024deadiff}  utilizes Q-Formers for feature disentanglement to establish a robust style-semantic separation framework. Nevertheless, these methods necessitate additional training or fine-tuning, substantially increasing computational complexity.

DiffStyle \cite{jeong2024DiffStyle} proposed a training-free approach that dynamically adapts h-space features during generation. StyleSSP \cite{xu2025stylessp} improves style transfer effectiveness by employing negative guidance and optimizing frequency-adjusted sampling initialization. InstantStyle-Plus \cite{wang2024instantstyleplus} integrating Tile ControlNet and gradient-based style guidance to improve content-semantic consistency. However, these methods fail to harness the powerful generative capabilities of diffusion models and exhibit limited adaptability, making it challenging to generate high-quality stylized images.


To tackle these challenges, we propose AttenST that fully exploits self-attention and cross-attention mechanisms to dynamically regulate feature interactions, achieving an optimal style-content balance.



\section{Background}

SDXL (Stable Diffusion XL) \cite{podell2023sdxl}, as a member of the Latent Diffusion Model (LDM) \cite{rombach2022LDM} family, utilizes a pre-trained variational autoencoder \cite{kingma2013VAE} $\mathcal{E}$ to map the input image $I$ to its latent representations $x=\mathcal{E}(I)$. The diffusion process operates entirely within this latent space, where the model learns to progressively denoise corrupted representations. The fundamental training objective focuses on minimizing the reconstruction error between predicted and ground truth noise patterns, thereby enabling robust image generation from noisy latent states. More specifically, the model is trained through a sequential noise perturbation process in latent space, where at each timestep $t$, it predicts the noise component conditioned on both the latent state $x_t$ and additional conditioning inputs $c$. This training process can be formally expressed as:

\begin{equation}
    \mathcal{L}=\mathbb{E}_{x_0,\epsilon,t}\left[\|\epsilon-\epsilon_\theta(x_t,t,c)\|_2^2\right],
\end{equation}
where $x_0$ denote the initial latent representation, $\epsilon$ is the noise sampled from a standard normal distribution $\mathcal{N}(0,1)$, and $\epsilon_\theta(x_t,t,c)$ represents the model's predicted noise. 




\section{Method}

Given a content image $X^c$ and a reference style image $X^s$, our goal is to transfer the style features of $X^s$ to the content image $X^c$, while preserving its structural and semantic information, ultimately generating a stylized image $X^{cs}$.  The pipeline of AttenST is shown in \cref{fig:framework}. In the following section, we provide a detailed explanation of the proposed method. We start with the style-guided self-attention mechanism (\cref{4.1}), followed by the style-preserving inversion (\cref{4.2}). Furthermore, we present the proposed content-aware AdaIN method (\cref{4.3}) and the dual-feature cross-attention mechanism (\cref{4.4}).

\subsection{Style-Guided Self-Attention} \label{4.1}

The cross-attention mechanism employs the $query$ from the image and the $(key,value)$ from the text, establishing a connection between the two modalities. This mechanism facilitates the integration of text-related attributes while maintaining the overall semantic integrity of the image. Building on this insight, we conceptualize style as a form of guidance and propose a style-guided self-attention mechanism (SG-SA) analogous to the cross-attention mechanism. Our analysis reveals that $query$ of a content image encodes its semantic structure and spatial layout, while $key$ and $value$ dictate the rendered visual elements. Consequently, substituting $key$ and $value$ of the content image with those of the style image within self-attention layers enables the alignment of style-related visual attributes with the content image. 


Specifically, the content and style images are invert through the inversion process to obtain their corresponding latent noise representations. During this process, the $key$ and $value$ of the style image are extracted at each timestep, which are essential for effectively representing the reference style. To introduce style features while preserving content information, the latent noise representation of the content image $X_T^c$ 
 is utilized as the initial noise for denoising. At each timestep $t$, style-guided self-attention is performed by retaining the $Q^c$ of the content image and replacing the $K^c$ and $V^c$ of the content image with $K^s$ and $V^s$ from the style image. By computing attention between $Q^c$, $K^s$, and $V^s$, style features are effectively integrated into the generation process, thus aligning content and style features. The proposed style-guided self-attention mechanism is formally defined as follows:
\begin{equation}
    \mathrm{Attention}(Q^c,K^s,V^s)=\mathrm{Softmax}(\frac{Q^cK^s{}^T}{\sqrt{d}})V^s,
\end{equation}
where $Q^c$ represents the query of the content image, $K^s$ and $V^s$ represent the key and value of the style image, respectively. $d$ denotes the dimension of $Q^c$.
\begin{figure}
    \centering
    \includegraphics[width=1\linewidth]{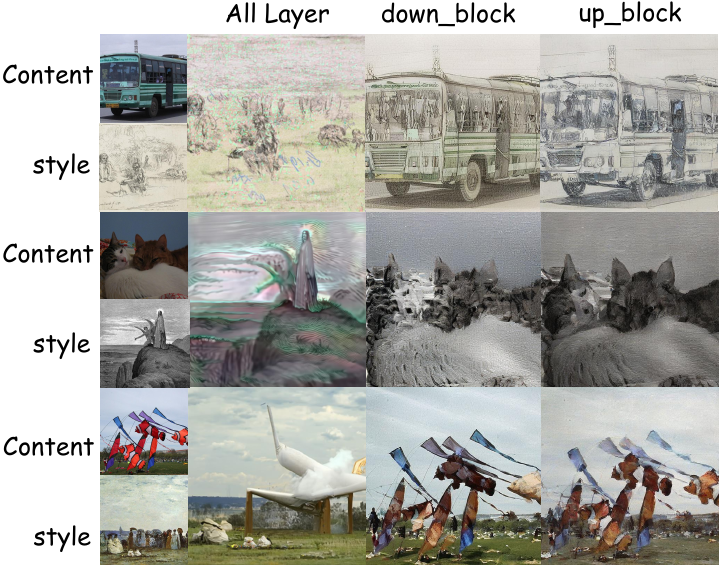}
    \caption{Qualitative results of style-guided self-attention mechanism application across different layers.}
    \label{fig:layer}
\end{figure}


However, simply applying the style-guided self-attention mechanism across all self-attention layers fails to achieve a balance between content and style. As illustrated in \cref{fig:layer}, injecting style features across all layers leads to noticeable style content leakage, while injecting them solely at the downsampling layers introduces some style information, yet significantly compromises the content structure. In contrast, incorporating style features into the upsampling module yields superior style transfer outcomes. Specifically, our experiments reveal that the optimal results are achieved by injecting style features at the $5^{th}$ and $6^{th}$ transformer blocks within the upsampling module, as further corroborated in \cref{5.5}.

\subsection{Style-Preserving Inversion}\label{4.2}
Most inversion-based style transfer methods rely on DDIM inversion \cite{song2020ddim}. Given a predefined timesteps $t=\{0,...,T\}$, the denoising process predicts the image $x_{t-1}$ from the current image $x_t$ based on \cref{eq4}. 

\begin{equation}
\begin{aligned}
        x_{t-1} &= \sqrt{\bar{\alpha}_{t-1}} \left( \frac{x_t - \sqrt{1 - \bar{\alpha}_t} \, \epsilon_\theta(x_t, t, c)}{\sqrt{\bar{\alpha}_t}} \right) \\ &+ \sqrt{1 - \bar{\alpha}_{t-1}} \, \epsilon_\theta(x_t, t, c)
\end{aligned}
\label{eq4}
\end{equation}
where $\bar{\alpha}_{t}$ denotes time-dependent noise schedules, and $\epsilon_\theta(x_t, t, c)$ represents the noise predicted by the UNet model under the text condition $c$.

The goal of DDIM inversion is to map the original image $x_0$ back to its corresponding noise representation $x_T$, generating a series of reverse trajectories $x_0,x_1...x_T$. Applying the inverse operation of \cref{eq4}, we formulate the inversion sampling equation \cref{eq5}, which establishes the mapping from $x_{t-1}$ to $x_{t}$. However, directly applying \cref{eq5} to solve $x_t$ is not feasible, as $\epsilon_\theta(x_t,t,c)$ depends on $x_{t}$. To solve this, DDIM relies on a linear assumption $\epsilon_\theta(x_t,t,c)\approx\epsilon_\theta(x_{t-1},t,c)$, which introduces errors that accumulate and ultimately degrade the style information during the inversion process.
\begin{equation}
\begin{aligned}
        x_t &=\sqrt{\frac{\bar{\alpha}_t}{\bar{\alpha}_{t-1}}}x_{t-1}+\sqrt{\bar{\alpha}_t}(\sqrt{\frac{1}{\bar{\alpha}_t}-1}
        \\ &-\sqrt{\frac{1}{\bar{\alpha}_{t-1}}-1})\epsilon_\theta(x_t,t,c)
\end{aligned}
\label{eq5}
\end{equation}

To address this issue, we propose a style-preserving inversion strategy that minimizes the loss of style information during the inversion process. At each iteration of the inversion, we perform multiple resampling steps to achieve more accurate reverse sampling.

\begin{figure}
    \centering
    \includegraphics[width=1\linewidth]{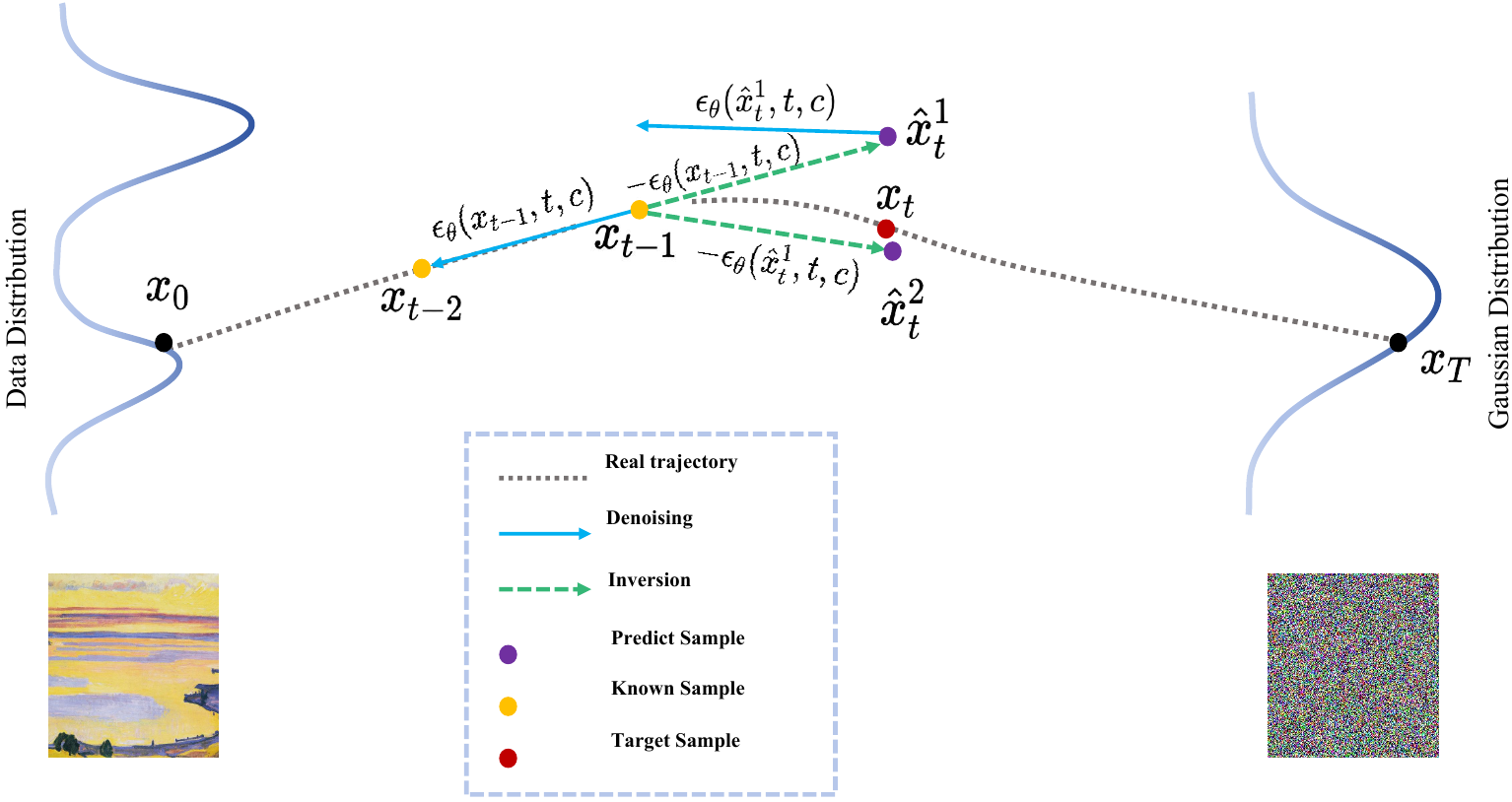}
    \caption{\textbf{Style-preserving inversion process.} We utilize the linear assumption to obtain $\hat{x}^1_{t}$, which provides a more accurate inversion direction compared to $x_{t-1}$. We then establish a refined inversion direction $-\epsilon_\theta(\hat{x}_t^1,t,c)$ by reversing the denoising trajectory from $\hat{x}^1_{t}$ to $x_{t-1}$, yielding a more precise estimated point $\hat{x}^2_{t}$.}
    \label{fig:inversion}
\end{figure}

As illustrated in \cref{fig:inversion}, consider the inversion from $x_{t-1}$ to $x_t$. For an ideal inversion, the trajectory from $x_{t-1}$ to $x_t$ should align with the denoising trajectory from $x_t$ to $x_{t-1}$. Unlike prior methods that approximate the inversion direction from $x_{t-1}$ to $x_{t}$ by reversing the denoising trajectory from $x_{t-1}$ to $x_{t-2}$, thereby yielding an approximate estimate $\hat{x}^1_{t}$, we find that $\hat{x}^1_{t}$ provides a more accurate direction than $x_{t-1}$. Thus, we retain $x_{t-1}$ as the starting point of the current inversion step and input the approximated $\hat{x}^1_{t}$ into the model for denoising ($\hat{x}^1_{t} \rightarrow x_{t-1}$). The reverse direction of this denoising step is subsequently utilized to refine the inversion trajectory from $x_{t-1}$ to $x_t$, producing a sample $\hat{x}^2_{t}$ that more closely approximates the target latent $x_t$. By repeating this resampling process for $n$ iterations, we obtain more precise reverse sampling points $\hat{x}^n_{t}$, effectively alleviating style loss during the inversion process.

\subsection{Content-Aware AdaIN} \label{4.3}
Adaptive instance normalization (AdaIN) \cite{huang2017AdaIN} is a style transfer technique that aligns the mean and variance of content features with those of style features. 
To effectively integrate style information at the early stage of generation, we manipulate the initial content noise using AdaIN. Given the latent noise representations of the content and style images, denoted as $x_{T}^{c}$ and $x_{T}^{s}$, respectively, the AdaIN operation is defined as follows:


\begin{equation}
    x_{T}^{cs}=\sigma(x_{T}^{s})\left(\frac{x_{T}^{c}-\mu(x_{T}^{c})}{\sigma(x_{T}^{c})}\right)+\mu(x_{T}^{s}),
\end{equation}\label{eq6}
where $\mu(\cdot)$ and $\sigma(\cdot)$ denote channel-wise mean and standard deviation.

While AdaIN successfully incorporates style features through feature statistic alignment, this process inevitably compromises the original content statistics, leading to substantial content degradation in the synthesized image. To this end, we propose a content-aware AdaIN method, which can be defined as follows:

\begin{equation}
\begin{aligned}
    x_{T}^{cs}=&(\alpha_s\sigma(x^s_T)+\alpha_c\sigma(x^c_T))\left(\frac{x-\mu(x^c_T)}{\sigma(x^c_T)}\right)
    \\ &+(\alpha_s\mu(x^s_T)+\alpha_c\mu(x^c_T))
\end{aligned}
\label{eq7}
\end{equation}
where $\alpha_c$ and $\alpha_s$ are parameters controlling the strength of the content and style features, and $\alpha_c+\alpha_s=1$.

The introduction of the content weight $\alpha_c$ enables CA-AdaIN to retain a portion of the content feature statistics during normalization. By adjusting the ratio of $\alpha_c$ and $\alpha_s$, CA-AdaIN dynamically balances the representation of content and style, effectively mitigating the loss of content information during the style transfer process.

\begin{table*}
\centering
\caption{Quantitative comparison with state-of-the-art methods. Lower values for all metrics indicate better performance.}
\label{tab:quantitative}
\renewcommand{\arraystretch}{1.3} 
\resizebox{1\textwidth}{!}{ 
\begin{tabular}{c|>{\columncolor{gray!20}}ccccccc|cccc}
\toprule
\textbf{Metrics} & \textbf{Ours} & \textbf{StyleID} \cite{chung2024styleID} & \textbf{DiffuseIT} \cite{kwon2022DiffuseIT} & \textbf{InST} \cite{zhang2023InST} & \textbf{DiffStyle} \cite{jeong2024DiffStyle}& \textbf{StyleAlign} \cite{hertz2024styleAlign}& \textbf{InstantStyle} \cite{wang2024instantstyle}& \textbf{AdaIN} \cite{huang2017AdaIN}& \textbf{AesPA-Net} \cite{hong2023aespa}& \textbf{AdaConv} \cite{chandran2021AdaConv}& \textbf{StyTR2} \cite{deng2022stytr2}\\ 
\midrule
ArtFID$\downarrow$  & \textbf{28.693}& 31.613& 41.965& 35.846& 42.486& 35.349& 38.596& 32.515& 32.080& 32.094& 30.419\\ 
FID$\downarrow$     & \textbf{18.559}& 20.190& 23.818& 20.068& 22.051& 20.827& 21.824& 19.337& 20.247& 19.294& 18.722\\ 
LPIPS$\downarrow$   & \textbf{0.467}& 0.492& 0.691& 0.702& 0.843& 0.620& 0.691& 0.599& 0.509& 0.581& 0.542\\ 
\bottomrule
\end{tabular}
}
\renewcommand{\arraystretch}{1.0} 
\end{table*}

\subsection{Dual-Feature Cross-Attention} \label{4.4}
Cross-attention serves as a fundamental mechanism for guiding image generation processes. To achieve enhanced style integration while preserving content fidelity, we propose a dual-feature cross-attention (DF-CA) mechanism. This innovative approach maximizes the potential of attention mechanisms by simultaneously embedding both content and style features into the generation process through cross-attention mechanism. Building upon the original SDXL framework, where text features $y$ interact with image query features ${I}$ as defined in \cref{eq8}:

\begin{equation}
    \phi^{text}=Attention({Q},{K},{V})=Softmax(\frac{{QK^T}}{\sqrt{d}}){V}
    \label{eq8}
\end{equation}
where ${Q}={I}{W}_q$, ${K}=\boldsymbol{y}{W}_k$,
${V}=\boldsymbol{y}{W}_v$ represent the value, query, and key matrices. ${W}_q$, ${W}_k$, ${W}_v$ are trainable weight matrices.

We employ pre-trained CLIP \cite{radford2021CLIP} image encoders to extract semantically-aligned feature embeddings from content and style images. These embeddings capture the intrinsic semantic relationships and visual characteristics, providing robust representations of both content structure and stylistic elements. Following this, we compute the cross-attention for the content and style features using \cref{eq9} and \cref{eq10}, respectively.

\begin{equation}
    \phi^c=Attention({Q},{K^c},{V^c})=Softmax(\frac{{Q{K^c}^T}}{\sqrt{d}}){V^c}
    \label{eq9}
\end{equation}
\begin{equation}
    \phi^s=Attention({Q},{K^s},{V^s})=Softmax(\frac{{Q{K^s}^T}}{\sqrt{d}}){V^s}
    \label{eq10}
\end{equation}
where ${Q}={I}{W}_q$, ${K^c}=\boldsymbol{c}{W}^{\prime}_k$,
${V^c}=\boldsymbol{c}{W}^{\prime}_v$, ${K^s}=\boldsymbol{s}{W}^{\prime}_k$,
${V^s}=\boldsymbol{s}{W}^{\prime}_v$.  $\boldsymbol{c}$ and $\boldsymbol{s}$ represent the content and style image feature respectively. ${W}^{\prime}_k$ and ${W}^{\prime}_v$ are pre-trained weight matrices \cite{ye2023Ip-adapter} for images.

The extracted image features are then integrated into the UNet via decoupled cross-attention. The final cross-attention calculation is demonstrated in the following equation:

\begin{equation}
    \phi^{final}= \phi^{text} +\phi^{c} +\phi^{s}
\end{equation}
The proposed DF-CA mechanism effectively integrates content and style features via cross-attention, enhancing the quality and reliability of the generated results.


\section{Experiments}
\subsection{Implementation Details}

\begin{figure*}
    \centering
    \includegraphics[width=1\linewidth]{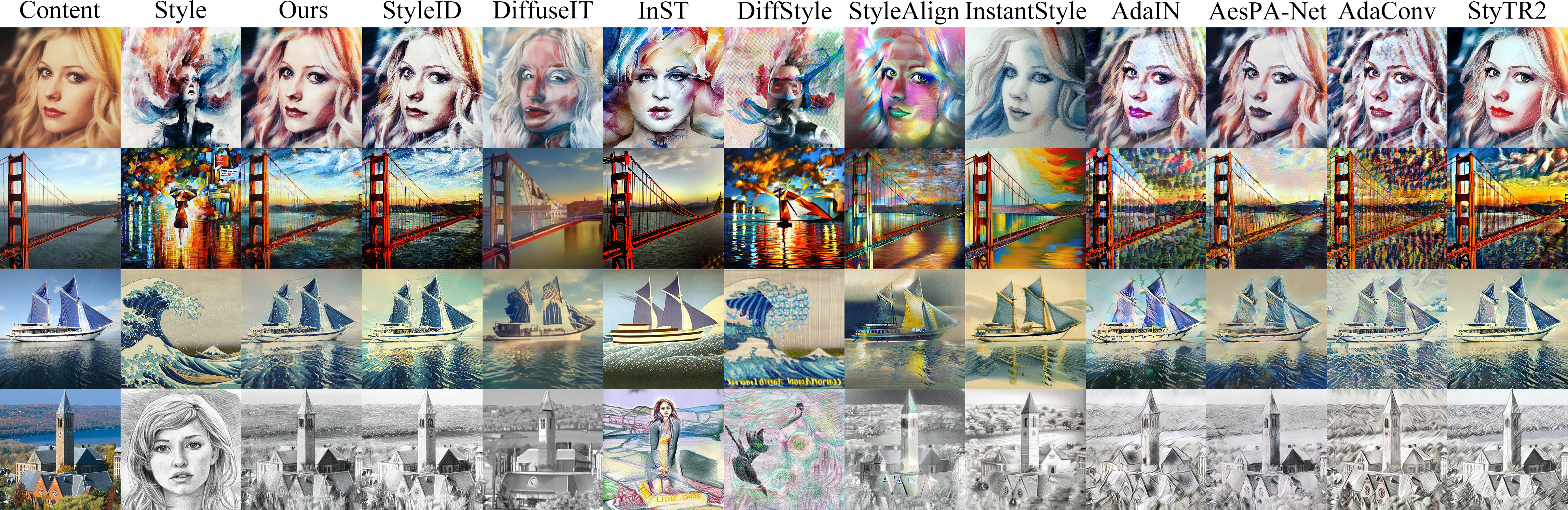}
    \caption{Qualitative comparison with state-of-the-art methods.}
    \label{fig:qualitative}
\end{figure*}

We conducted experiments in the SDXL v1.0 base model, with both the sampling and inversion steps configured to 20. The number of resampling steps for SPI was set to 5. For the DF-CA, the cross-attention features corresponding to content and style were injected into the last downsampling and first upsampling transformer blocks respectively. Text prompts for images were produced using the BLIP2 \cite{li2023blip} model.
The weighting parameters $\alpha_c$ and $\alpha_s$ were set to 0.4 and 0.6. All experiments were executed on a single NVIDIA 3090 GPU with 24GB of memory.

\textbf{Dataset}. The content and style images utilized for evaluation were sourced from the MS-COCO \cite{lin2014mscoco} and WikiArt \cite{tan2018wikiart} datasets. For a fair comparison, 20 content images and 40 style images were randomly selected, with all input images resized to a resolution of 512 × 512. Ultimately,  800 stylized images are obtained.

\textbf{Evaluation Metrics}. We utilized three established metrics: FID \cite{heusel2017FID}, LPIPS \cite{zhang2018LPIPS}, and ArtFID \cite{wright2022artfid}. FID quantifies the similarity between stylized images and reference style images. LPIPS evaluates perceptual differences in structure and texture between stylized and content images. ArtFID integrates content and style fidelity, providing a comprehensive evaluation aligned with human preference. ArtFID is computed as $ArtFID=(1+LPIPS) \cdot (1+FID)$.

\subsection{Quantitative Results}
In this section, we conduct a comparative analysis of AttenST against state-of-the-art methods, encompassing both traditional approaches (AdaIN \cite{huang2017AdaIN}, AesPA-Net \cite{hong2023aespa},  StyTR2 \cite{deng2022stytr2}, AdaConv \cite{chandran2021AdaConv}, CAST \cite{zhang2022CAST}, EFDM \cite{zhang2022EFDM}, MAST \cite{deng2020MAST}, AdaAttn \cite{liu2021adaattn}, ArtFlow \cite{an2021artflow}) and diffusion-based methods (StyleID \cite{chung2024styleID}, DiffuseIT \cite{kwon2022DiffuseIT}, InST \cite{zhang2023InST}, DiffStyle \cite{jeong2024DiffStyle}, StyleAlign \cite{hertz2024styleAlign}, InstantStyle \cite{wang2024instantstyle}). Due to space limitations, the results for CAST, EFDM, MAST, AdaAttn, and ArtFlow are included in the appendix. 



As illustrated in \cref{tab:quantitative}, our method consistently surpasses all comparative approaches across every evaluation metric. 
Specifically, our method achieves an FID score of 18.559, demonstrating a significant improvement over both DiffuseIT and DiffStyle, which underscores its advanced style fusion capabilities. The superior performance is further evidenced by the optimal LPIPS score of our method, indicating exceptional perceptual similarity preservation. While AdaIN shows a comparable FID score, its substantially higher LPIPS value reveals significant structural and textural degradation. Furthermore, our method achieves a best ArtFID score of 28.693, confirming its effectiveness in transferring the desired style while preserving the integrity of the content.

\subsection{Qualitative Results}
As shown in \cref{fig:qualitative}, our method demonstrates superior image quality and style transfer performance compared to existing approaches. The qualitative analysis reveals several key observations: (1) Diffusion-based methods exhibit varying limitations: DiffStyle suffers from style content leakage, StyleAlign and InstantStyle show insufficient detail retention despite ControlNet implementation, InST achieves partial content preservation but lacks style consistency. Although StyleID achieves comparable visual quality, it suffers from elevated luminance levels and compromised content fidelity.  (2) Traditional methods, while demonstrating basic style transfer capabilities, are prone to produce noticeable artifacts, particularly in AdaConv and AdaIN implementations. In contrast, our approach achieves significantly better performance across both style transfer fidelity and visual quality, effectively overcoming the limitations inherent in existing approaches while maintaining computational efficiency.


\subsection{Ablation Study}
\begin{figure}
    \centering
    \includegraphics[width=1\linewidth]{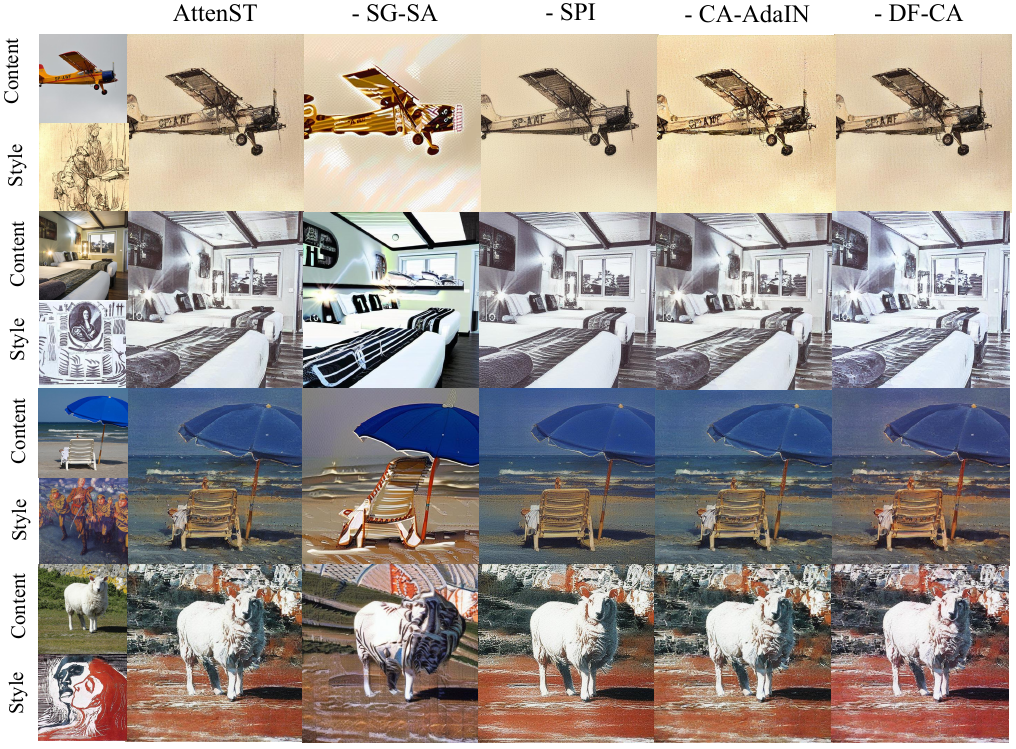}
    \caption{Qualitative ablation study of our method.}
    \label{fig:ablation_imgs}
\end{figure}

In this section, we conduct comprehensive ablation studies to systematically evaluate the contribution of each component in our framework. As detailed in \cref{tab:ablation} and \cref{fig:ablation_imgs}, we investigate four distinct configurations corresponding to the methodological components presented in \cref{4.1}-\cref{4.4}: (1) - SG-SA: removal of the style-guided self-attention mechanism; (2) - SPI: replacement of our style-preserving inversion with standard DDIM inversion; (3) - CA-AdaIN: substitution of our content-aware AdaIN with original AdaIN; and (4) - DF-CA: elimination of the dual-feature cross-attention mechanism. 

\begin{table}
\centering
\caption{Ablation study of each component of our method.}
\label{tab:ablation}
\begin{tabular}{lccc}
\hline
Method           & ArtFID$\downarrow$           & FID$\downarrow$        & LPIPS$\downarrow$           \\ \hline
\rowcolor{gray!20}
AttenST             & \textbf{28.693}& \textbf{18.559}& \textbf{0.467}\\
- SG-SA& 40.133& 25.839& 0.495\\
- SPI& 36.448& 24.107& 0.452\\
- CA-AdaIN        & 31.093& 19.772& 0.497\\
- DF-CA& 31.906& 20.500& 0.484\\ \hline
\end{tabular}
\end{table}


Experimental results show that the absence of the style-guided self-attention mechanism leads to a significant deterioration in the FID score, highlighting its critical role in achieving effective style fusion. Moreover, the use of DDIM inversion results in substantial style information loss, demonstrating its limitations in preserving stylistic attributes. Notably, CA-AdaIN outperforms traditional AdaIN by seamlessly integrating style information while maintaining content fidelity. Additionally, DF-CA proves indispensable in balancing content preservation and style transfer through its advanced cross-attention mechanisms. Taken together, these findings collectively validate the efficacy and necessity of each proposed component within our framework.

\subsection{Content-Style Trade-off Analysis} \label{5.5}
\begin{figure}
  \centering
  \begin{subfigure}{0.48\linewidth}
    \includegraphics[width=\linewidth]{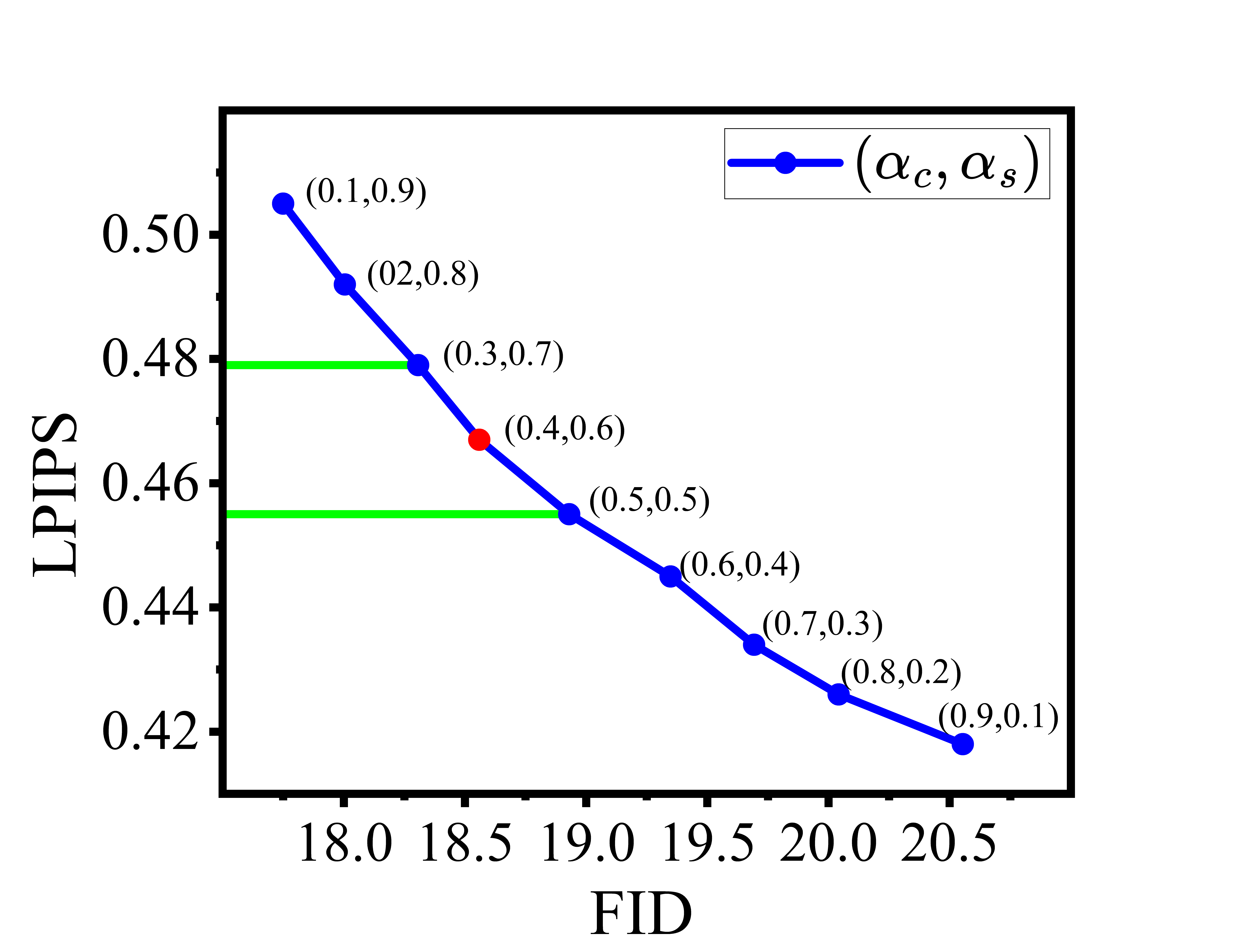}
    \caption{Comparison between $\alpha_c$ and $\alpha_s$}
    \label{fig:a}
  \end{subfigure}
  \hfill
  \begin{subfigure}{0.48\linewidth}
    \includegraphics[width=\linewidth]{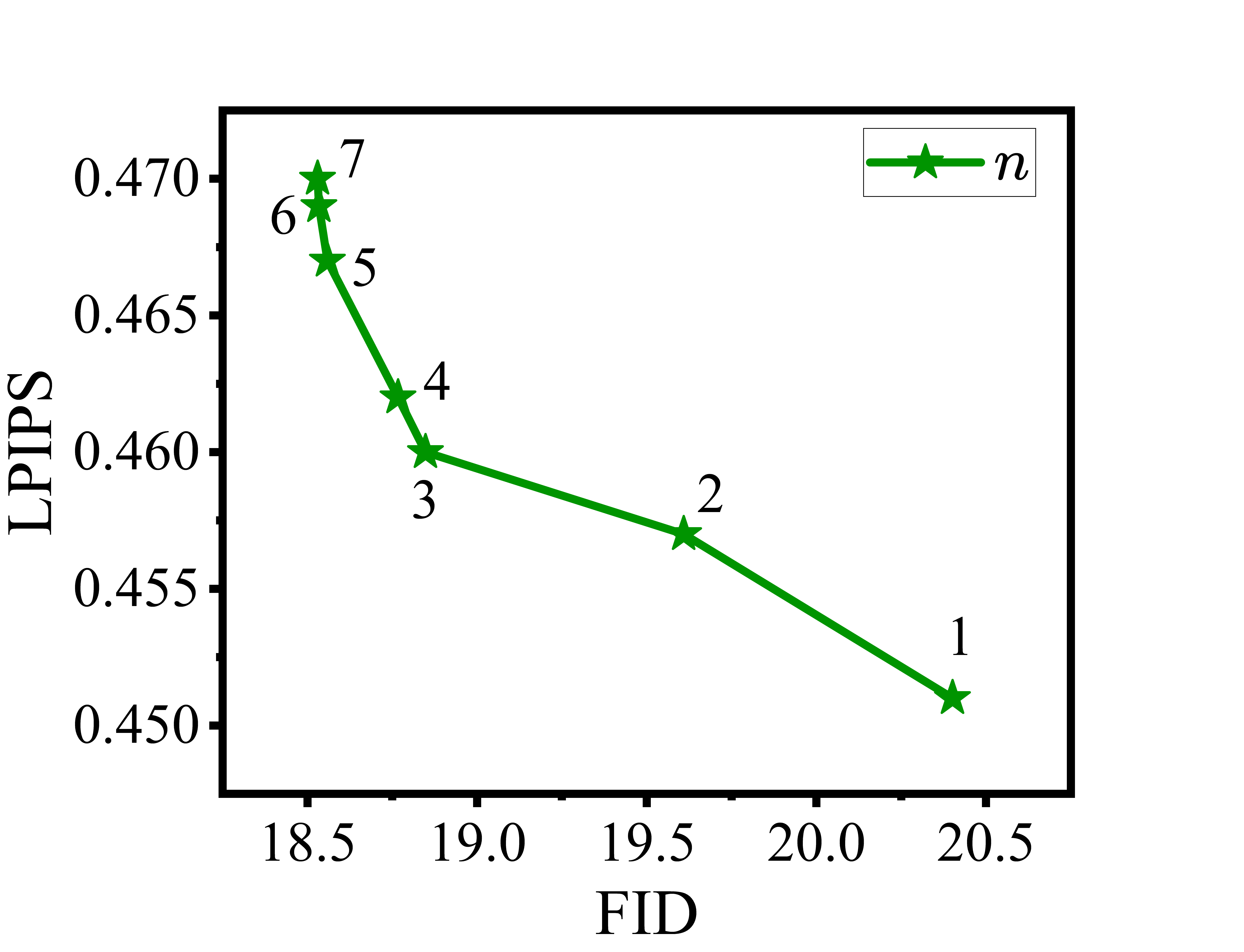}
    \caption{Comparison across values of $n$}
    \label{fig:b}
  \end{subfigure}
  \caption{Impact of $\alpha_c$ , $\alpha_s$, and $n$ on style transfer results.}
  \label{fig:parameter}
\end{figure}

\begin{figure}
    \centering
    \includegraphics[width=1\linewidth]{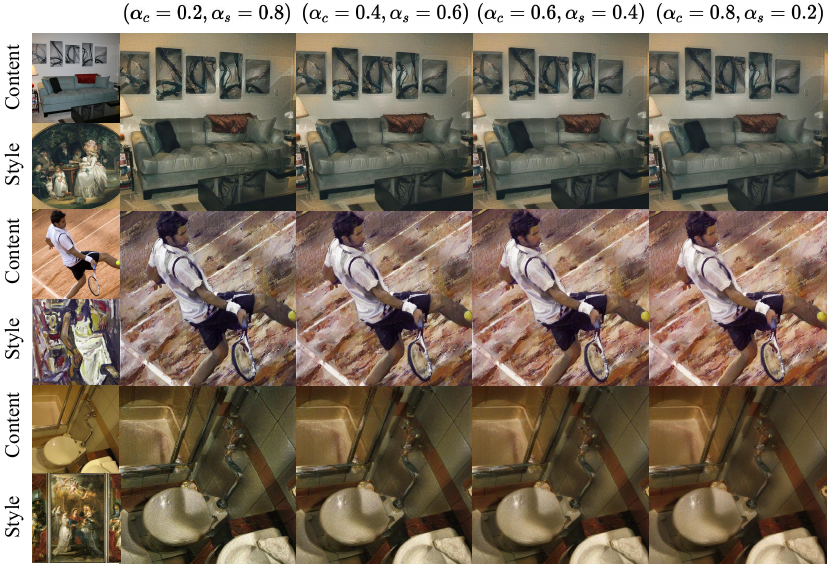}
    \caption{Visualization of the effects of different values of $\alpha_c$ and $\alpha_s$. }
    \label{fig:viusal-weight}
\end{figure}

Proposed CA-AdaIN incorporates $\alpha_s$ and $\alpha_c$ that enable flexible control over the strength of style transfer and content preservation. As illustrated in \cref{fig:a} and \cref{fig:viusal-weight}, as the parameter $\alpha_c$ increases, the LPIPS metric gradually declines, while the FID score shows a corresponding increase. This correlation indicates enhanced content preservation at the expense of reduced style integration.
Our analysis reveals that LPIPS values within the range of 0.45-0.48 (green lines in \cref{fig:a}) represent an optimal balance between content preservation and style transfer. 
Consequently, we establish $(\alpha_c=0.4, \alpha_s=0.6)$ as the default parameter configuration, providing an effective trade-off between stylistic expression and content fidelity.

\textbf{Study on resampling steps $\boldsymbol{n}$}. We further investigated the impact of the resampling steps $n$ of the style-preserving inversion. As shown in \cref{fig:b}, increasing $n$ leads to a corresponding decrease in the FID score, albeit at the cost of a minor content loss. However, as $n$ continues to grow, the reduction in FID plateaus, indicating that the refined inversion trajectory has become sufficiently close to the target inversion trajectory. Beyond this point, further resample steps no longer significantly enhances style preservation. Consequently, we selected $n = 5$ as the optimal resampling step. 

\begin{figure}
    \centering
    \includegraphics[width=0.9\linewidth]{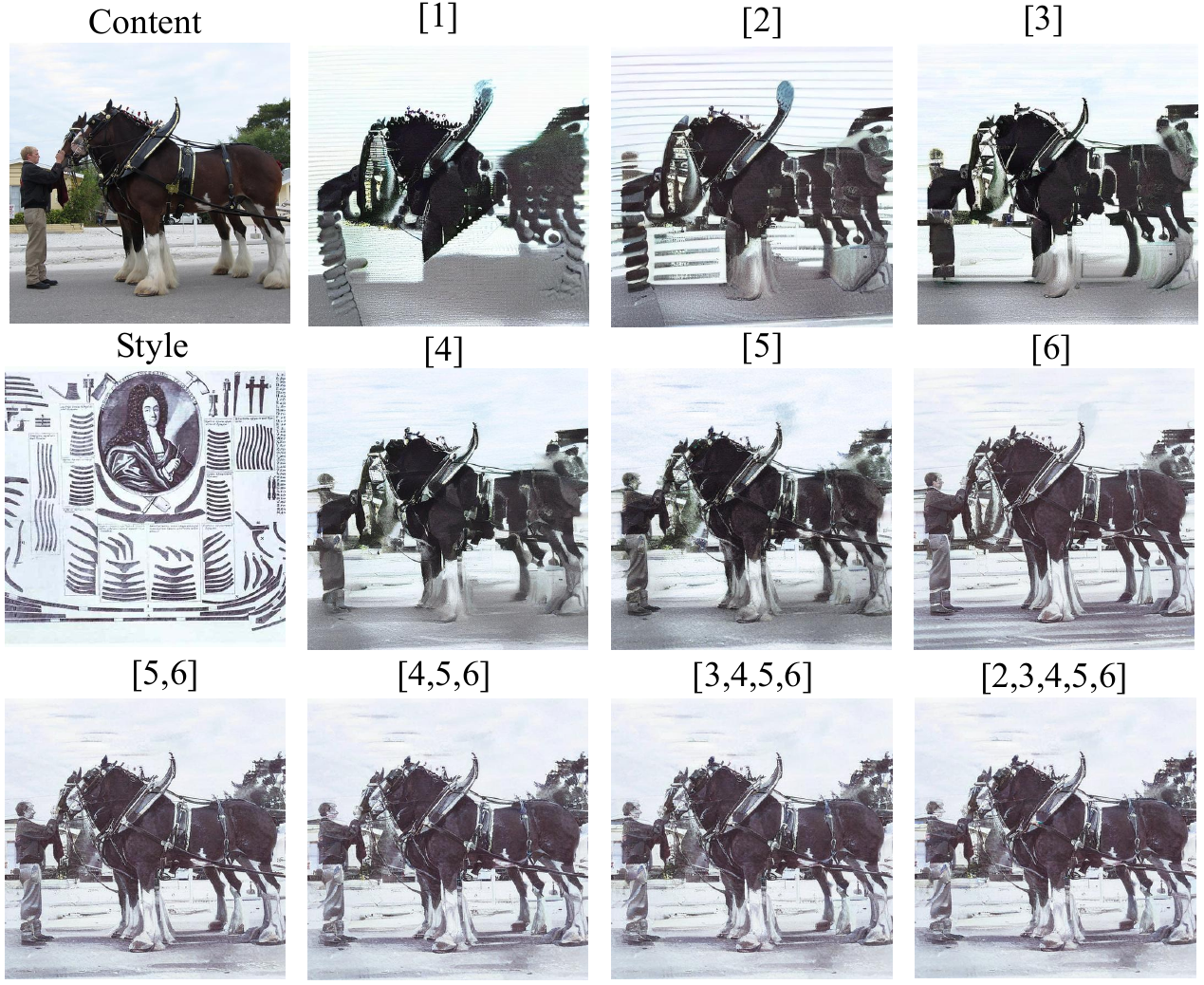}
    \caption{Visualization of style injection effects across different blocks.}
    \label{fig:layer_ablation}
\end{figure}

\textbf{Analysis of style injection Layer}. As discussed in \cref{4.1}, we established the upsampling module as the optimal stage for style-guided self-attention, we conducted systematic experiments across the six transformer blocks within this module to identify the most effective injection positions.  As illustrated in \cref{fig:layer_ablation} and \cref{tab:layer}, injecting style in the early transformer blocks ($1^{th}$-$4^{th}$) not only fails to effectively integrate style but also leads to significant content distortion. In contrast, late-stage injection achieves a better balance between content preservation and style integration. Based on these findings, we further explored different combinations of late-stage blocks and identified that injecting style into the $5^{th}$ and $6^{th}$  transformer blocks yields the best results.

\begin{table}
\centering
\caption{Comparison of style injection results across different blocks.}
\label{tab:layer}
\begin{tabular}{cccc}
\hline
Injection Blocks& ArtFID$\downarrow$& FID$\downarrow$    & LPIPS$\downarrow$ \\ \hline
$1$
& 36.736 & 22.295 & 0.577 \\
$2$
& 34.138 & 21.182 & 0.539 \\
$3$
& 35.031 & 22.622 & 0.483 \\
$4$
& 32.304 & 20.827 & 0.48  \\
$5$
& 32.200 & 21.268 & 0.446 \\
$6$
& 30.267 & 19.731 & \textbf{0.460}  \\
\rowcolor{gray!20}
$[5,6]$
& \textbf{28.693} & 18.559 & 0.467 \\
$[4,5,6]$
& 28.775 & 18.439 & 0.492 \\
$[3,4,5,6]$
& 29.197 & \textbf{18.286} & 0.502 \\
$[2,3,4,5,6]$& 29.376 & 18.467 & 0.509 \\ \hline
\end{tabular}
\end{table}
\section{Conclusion}
In this work, we delve into the potential of attention mechanisms in diffusion models for style transfer tasks, introducing AttenST, a training-free attention-driven style transfer framework. Our approach achieves style infusion through strategic control of self-attention mechanisms, complemented by a style-preserving inversion method that significantly mitigates style loss during the inversion process. 
We further propose CA-AdaIN, which adaptively adjusts the initial noise while effectively preserving content information during style fusion. Furthermore, we introduce the DF-CA mechanism, which enables precise control over both content and style features in the generated images through a cross-attention mechanism. Extensive experimental results demonstrate that AttenST outperforms both conventional and diffusion-based methods, achieving state-of-the-art performance on style transfer dataset.

{
    \small
    \bibliographystyle{ieeenat_fullname}
    \bibliography{main}
}

\clearpage
\setcounter{page}{1}
\maketitlesupplementary

\section{Appendix}
\subsection{Details of Feature Preserving Inversion}
In this work, we propose a Feature Preserving Inversion (FPI) strategy to invert an input image $x_0$ into a latent noise representation $X_T$ while generating an inversion trajectory $\{x_t\}_{t=1}^{T}$. As shown in the \cref{alg:FPI}, the algorithm operates iteratively over time steps $t=1,2,\ldots,T$, with each step involving a refinement process to ensure the preservation of key image features. Given the input image $x_0$, time step $T$, and refinement iterations $n$, the algorithm outputs the latent representation $X_T$ and the trajectory $\{x_t\}_{t=1}^{T}$. 

\begin{algorithm}[!ht]
    \caption{Feature Preserving Inversion}
    \label{alg:FPI}
    \renewcommand{\algorithmicrequire}{\textbf{Input:}}
    \renewcommand{\algorithmicensure}{\textbf{Output:}}
    
    \begin{algorithmic}[1]
        \REQUIRE Image $x_0$,time step $T$,and refine iteration $n$.   
        \ENSURE Latent noise representation $X_T$ and inversion trajectory $\{x_t\}_{t=1}^{T}$.    
        
        \FOR{$t=1,2,\ldots,T$}
            \STATE $\hat{x}_t^{1}\leftarrow inversion(x_{t-1},t,c)$
            \FOR{$i=1,2,\ldots,n$}
                \STATE $\hat{x}_t^{i}\leftarrow inversion(\hat{x}^{i-1}_{t},t,c)$
            \ENDFOR
        \ENDFOR
        \RETURN ($X_T$, $\{x_t\}_{t=1}^{T}$)

        \STATE   
        \STATE \textbf{Function} $inversion(x_t, t,c)$:
        \STATE \quad $\begin{aligned}\hat{x}&=\sqrt{\frac{\bar{\alpha}_{t}}{\bar{\alpha}_{t-1}}}x_{t-1}-\sqrt{\bar{\alpha}_{t}}(\sqrt{\frac{1}{\bar{\alpha}_{t}}-1}\\&-\sqrt{\frac{1}{\bar{\alpha}_{t-1}}-1})\epsilon_\theta(x_t,t,c)\end{aligned}$ \quad 
        \STATE \quad \RETURN $\hat{x}$

    \end{algorithmic}
\end{algorithm}

\subsection{Text Prompts on Style Transfer}

\begin{figure}[!ht]
    \centering
    \includegraphics[width=1\linewidth]{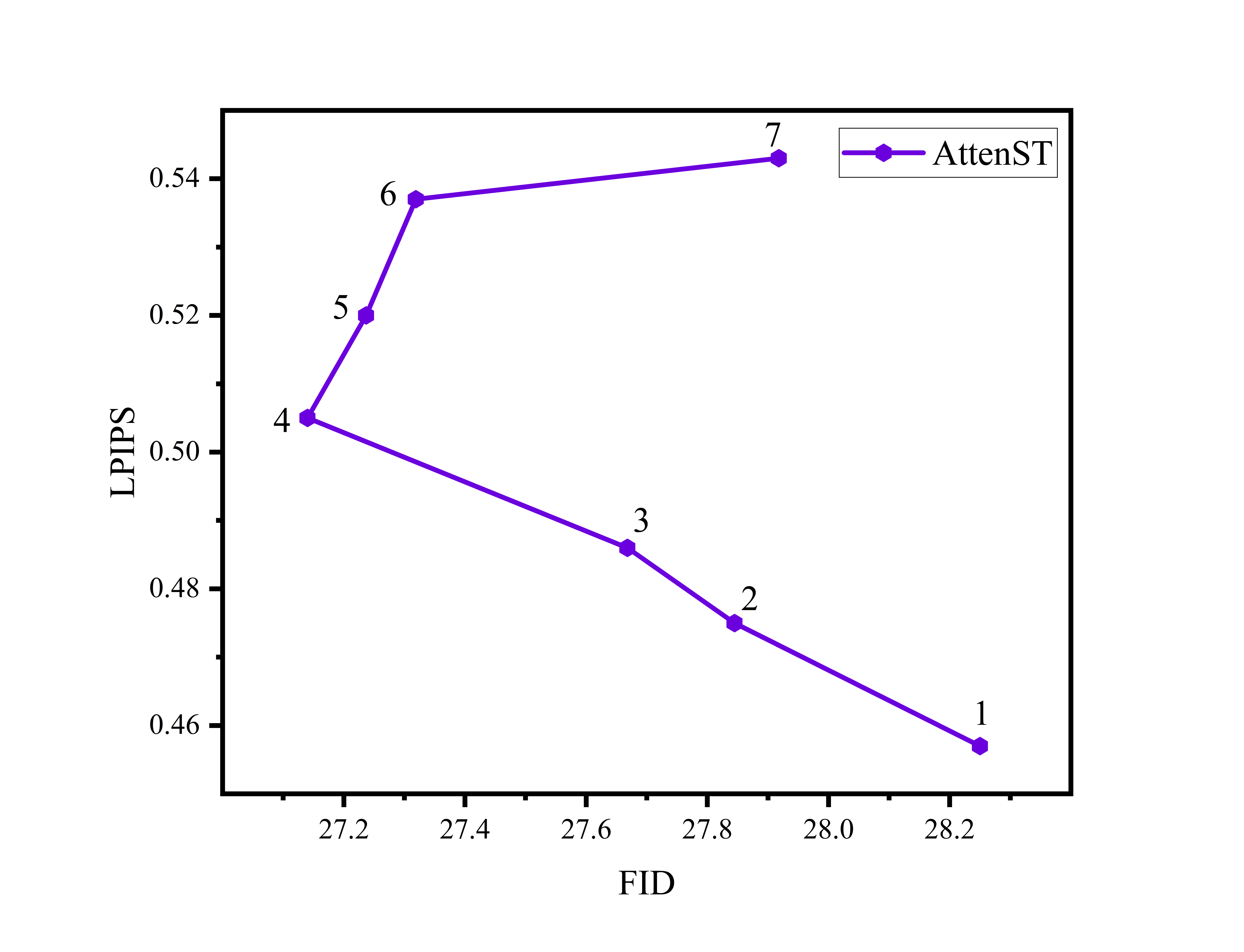}
    \caption{The impact of text prompts on style transfer results}
    \label{fig:cfg}
\end{figure}

\begin{figure*}[!ht]
    \centering
    \includegraphics[width=1\linewidth]{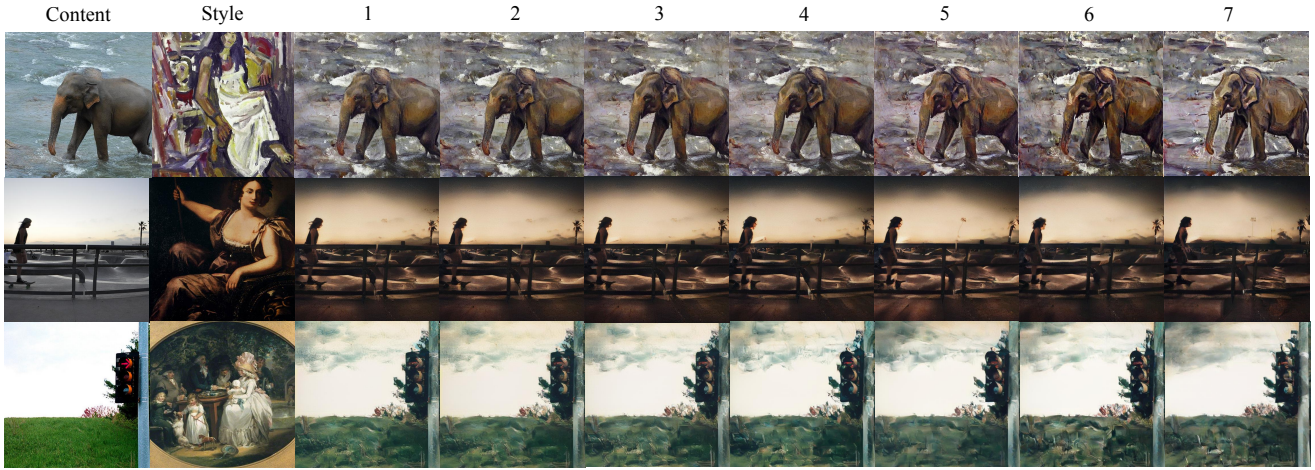}
    \caption{Visualization of results under different CFG guidance scales}
    \label{fig:cfg_vis}
\end{figure*}

To investigate the impact of text guidance on AttenST, we utilized guidance texts generated by the BLIP2 \cite{li2023blip} model to condition both the inversion and inference stages. Additionally, we examined the effect of different CFG guidance scale values. We selected ten content images and ten style images, generating a total of 100 stylized results for evaluation.  As shown in \cref{fig:cfg} and \cref{fig:cfg_vis}, increasing the CFG guidance scale enhances the style transfer effect; however, it also leads to a progressive degradation of the structural information in the content image. When the CFG guidance scale exceeds 4, both style transfer quality and content preservation continue to deteriorate, indicating that a larger CFG guidance scale does not necessarily yield better results. Excessively high CFG guidance scale values can negatively affect style transfer performance. Therefore, we set the default CFG guidance scale to 3 as a balanced choice.

\begin{figure*}[!ht]
    \centering
    \includegraphics[width=1\linewidth]{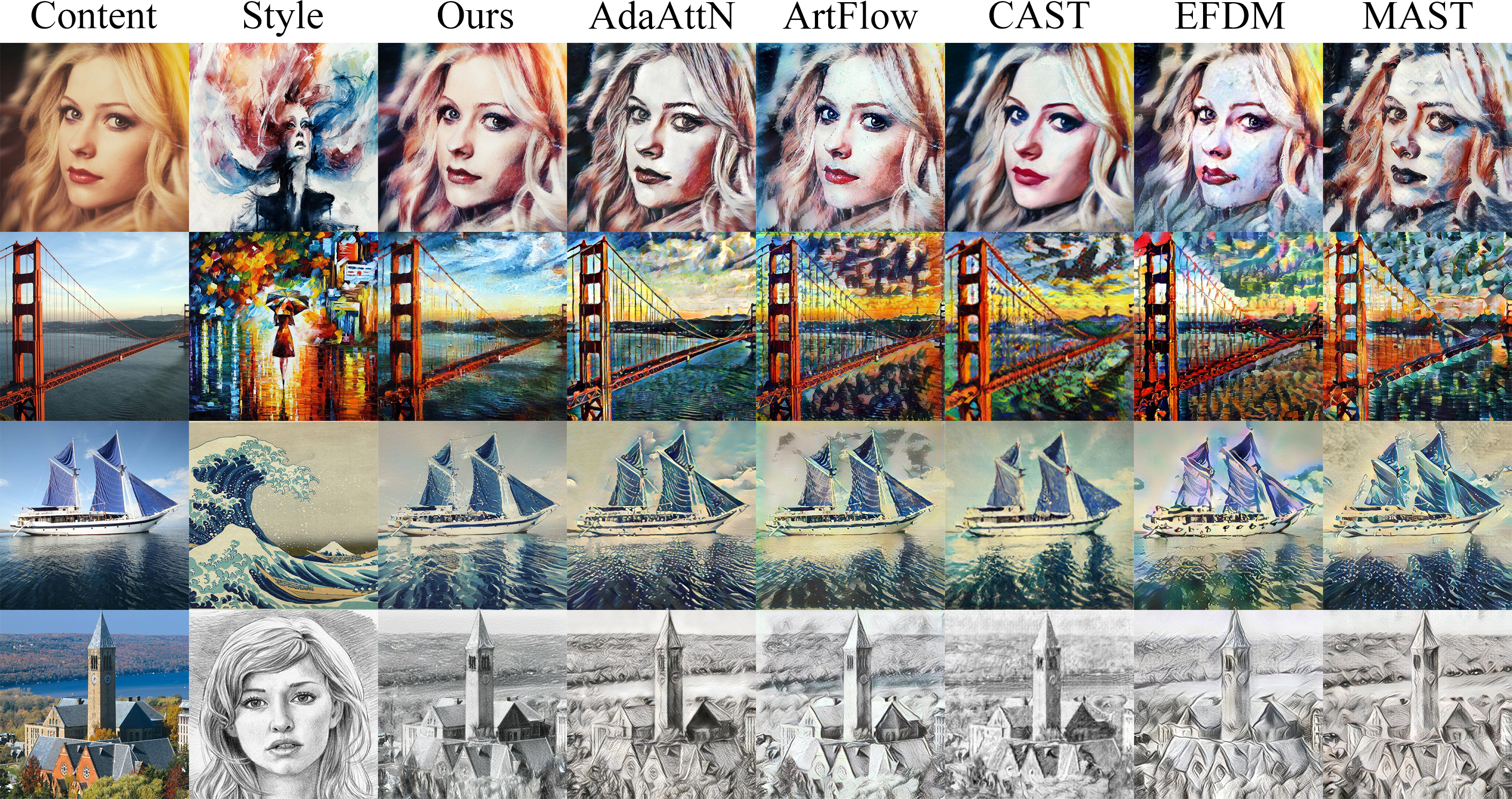}
    \caption{Additional results of traditional methods}
    \label{fig:additionanl_tra}
\end{figure*}

\subsection{Inference Efficiency}

\begin{table}[!ht]
\centering
\caption{Inference time comparison of diffusion-based style transfer methods for single-Image stylization.}
\label{tab:time}
\begin{tabular}{ccccc}
\hline
Metrics & Ours  & DiffuseIT & InST   & StyleID \\ \hline
Time(s) & 21.84 & 61.23& 824.29 & 15.86   \\ \hline
\end{tabular}
\end{table}

We compared the inference efficiency of AttenST with several diffusion-based style transfer methods. As shown in \cref{tab:time}, we performed style transfer on a single image using our method and three other approaches—DiffuseIT, InST, and StyleID—on an NVIDIA 3090 GPU, and measured the corresponding runtime. The results indicate that our method achieves significantly higher inference efficiency compared to DiffuseIT, and InST, while maintaining comparable efficiency to StyleID. This demonstrates the computational efficiency of our approach.

\subsection{User Study}

To further evaluate the perceptual quality of the stylized results generated by different methods, we conducted a comprehensive user study comparing our proposed AttenST with four state-of-the-art style transfer methods: StyTR2, StyleID, AdaIN, and AesPA-Net. The study aimed to assess the subjective preferences of human observers regarding the visual quality, style consistency, and content preservation of the stylized images.We randomly selected 10 content-style pairs, with each pair processed by all five methods. The study involved 30 participants. Each participant was presented with the stylized results in a randomized order and asked to rate the images based on the following criteria:
\begin{itemize}
    \item \textbf{Style Consistency: }How well the stylized image reflects the artistic style of the reference style image.
    \item \textbf{Content Preservation: }How well the original content structure and details are preserved in the stylized image.
    \item \textbf{Overall Visual Quality: }The overall aesthetic appeal and naturalness of the stylized image.
\end{itemize}

\begin{table}[!ht]
\centering
\renewcommand{\arraystretch}{1.2} 
\caption{User study results comparing AttenST with state-of-the-art style transfer methods. Scores are averaged across all participants and images.}
\label{tab:user_study}
\begin{tabularx}{\linewidth}{XXXX}
\hline
\textbf{Method} & \textbf{Style} & \textbf{Cotent} & \textbf{Overall} \\ \hline
\rowcolor{gray!20}
AttenST & 4.32 & 4.15 & 4.28 \\ 
StyTR2 & 3.89 & 3.54 & 3.64 \\ 
StyleID & 4.09 & 3.89 & 3.93 \\ 
AdaIN & 3.73 & 3.45 & 3.61 \\ 
AesPA-Net & 3.91 & 3.68 & 3.77 \\ \hline
\end{tabularx}
\end{table}

Participants rated each criterion on a 5-point Likert scale (1 = Poor, 5 = Excellent). The results were averaged across all participants and images for each method. The results of the user study are summarized in \cref{tab:user_study}. Our proposed AttenST consistently outperformed the competing methods across all three criteria. Specifically, AttenST achieved the highest scores in Style Consistency (4.32) and Overall Visual Quality (4.28), demonstrating its ability to effectively transfer artistic styles while maintaining high perceptual quality. In terms of Content Preservation, AttenST also ranked first with a score of 4.15, indicating its superior capability to retain the structural details of the original content. Compared to AesPA-Net and StyleID, which achieved moderate scores, AttenST showed significant improvements in both style transfer fidelity and content preservation. While AdaIN and StyTR2 performed reasonably well in terms of style consistency, they struggled to preserve fine-grained content details, resulting in lower scores for content preservation and overall visual quality.

\subsection{limitations}
Although the proposed AttenST effectively balances style integration and content preservation, enabling the generation of high-quality stylized results, certain limitations remain.  AttenST relies on a pre-trained diffusion model, making the quality of style transfer dependent on the priors learned by the base model. If the base model's feature representations are inadequate, the fidelity of the transferred style may be compromised.  When the reference style is highly abstract, the model may struggle to capture all stylistic elements and features. Although our method significantly reduces computational costs compared to optimization-based approaches, it still requires multiple diffusion steps to achieve high-quality results. Future work could explore integrating adaptive noise scheduling or adopting lightweight model variants to enhance inference efficiency.

\subsection{Additional qualitative results.}

\begin{figure*}[!ht]
    \centering
    \includegraphics[width=1\linewidth]{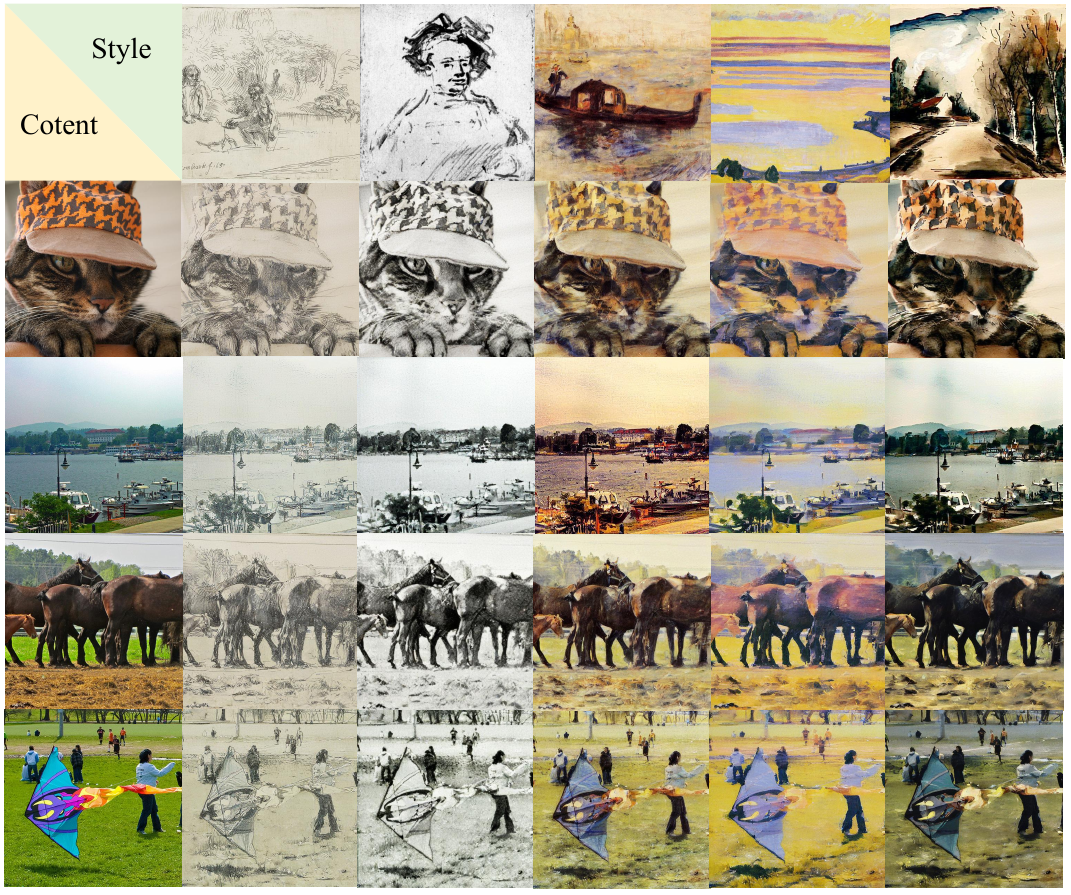}
    \caption{Additional results of AttenST}
    \label{fig:additionanl_tra2}
\end{figure*}

\begin{figure*}
    \centering
    \includegraphics[width=1\linewidth]{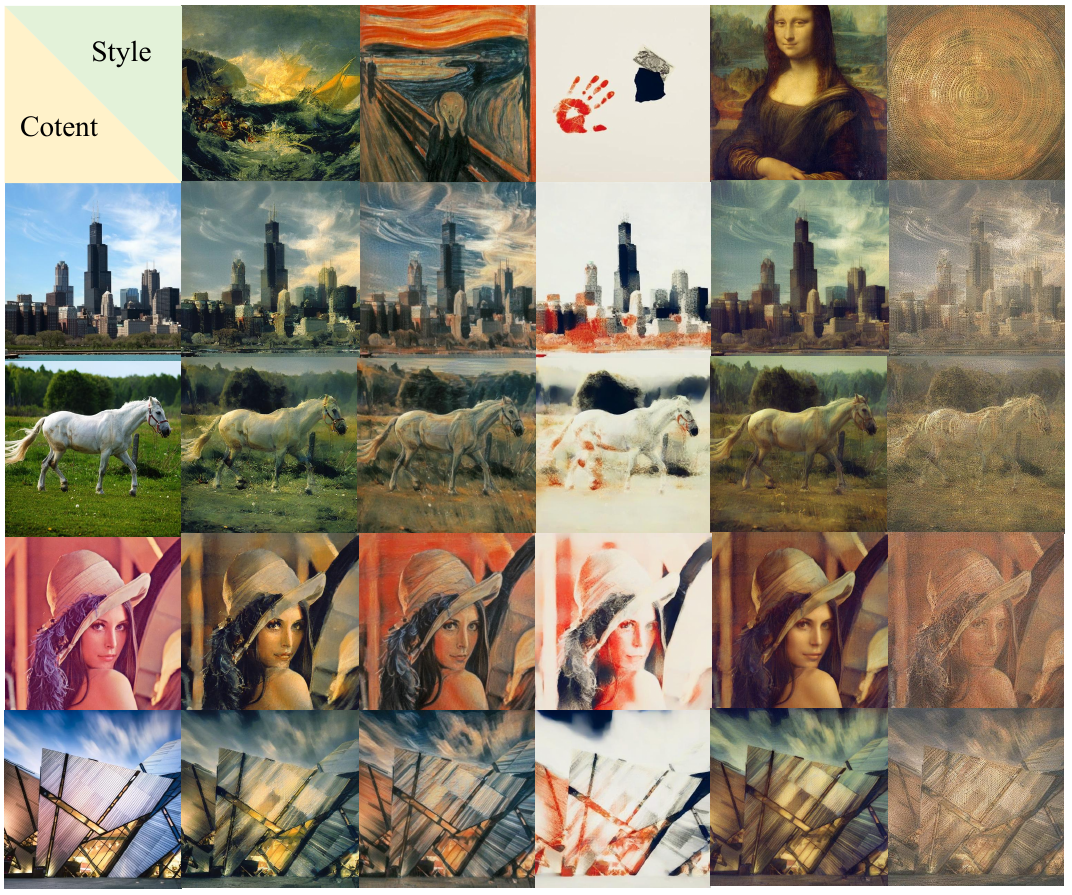}
    \caption{Additional results of AttenST}
    \label{fig:additionanl_tra3}
\end{figure*}

We further present additional qualitative comparison results. \Cref{fig:additionanl_tra} illustrates a qualitative comparison between our method and five traditional approaches: AdaAttn, ArtFlow, CAST, EFDM, and MAST. \Cref{fig:additionanl_tra2} and \cref{fig:additionanl_tra3} showcase the qualitative results of AttenST across different content-style image pairs.

\end{document}